\documentclass[lettersize,journal]{IEEEtran}
\usepackage{amsmath,amsfonts}
\usepackage{algorithmic}
\usepackage{algorithm}
\usepackage{array}
\usepackage[caption=false,font=normalsize,labelfont=sf,textfont=sf]{subfig}
\usepackage{textcomp}
\usepackage{stfloats}
\usepackage{url}
\usepackage{verbatim}
\usepackage{graphicx}
\usepackage{cite}
\usepackage{xcolor}
\graphicspath{{./Figures/}}

\begin{document}

\title{Design and Control of a High-Performance Hopping Robot}

\author{Samuel Burns,
        Matthew Woodward
\thanks{Manuscript received December 30, 2024; revised.} 
\thanks{The authors are with the Robot Locomotion and Biomechanics Laboratory, Mechanical Engineering Department, Tufts University, Medford, MA, 01890 (email: matthew.woodward@tufts.edu)}}



\maketitle

\begin{abstract}
Jumping and hopping locomotion are efficient means of traversing unstructured rugged terrain with the former being the focus of roboticists; a focus that has recently been changing. This focus has led to significant performance and understanding in jumping robots but with limited practical applications as they require significant time between jumps to store energy, thus relegating jumping to a secondary role in locomotion. Hopping locomotion, however, can preserve and transfer energy to subsequent hops without long energy storage periods. However, incorporating the performance observed in jumping systems into their hopping counterparts is an ongoing challenge. To date, hopping robots typically operate around 1 meter with a maximum of 1.63 m whereas jumping robots have reached heights of 30 m. This is due to the added design and control complexity inherent in developing a system able to input and store the necessary energy while withstanding the forces involved and managing the system's state. Here we report hopping robot design principles for efficient, robust, high-specific energy, and high-energy input systems through analytical, simulation, and experimental results. The resulting robot (MultiMo-MHR) can hop over 4 meters ($\sim$2.4x the current state-of-the-art) and is designed to withstand impact at terminal velocity ($\geq 30.7$ m).
\end{abstract}

\begin{IEEEkeywords}
Hopping, Robot, Control, Energy Efficiency
\end{IEEEkeywords}

\section{Introduction}
\IEEEPARstart{J}{umping} robots have been under significant development for decades, demonstrating a diversity of designs, energy storage and release mechanisms, and additional behaviors such as gliding and self righting \cite{Armour2007,Zhang2020}. Jumping locomotion tends to produce the highest specific energy (energy/mass) systems \cite{Hawkes2022,BostonDynamics2013}, as compared to hopping because they do not require the additional components necessary for hopping locomotion; namely those for orientation control. The challenge becomes how to incorporate that level of performance into hopping robots.

Robot platforms that perform single jumps are more common than those that hop continuously. However, due to the similarities in behavior, it is beneficial to apply findings from jumping platforms to their hopping counterparts. Jumping platforms have outperformed hopping systems in terms of energy storage and specific energy, with high-performance jumping platforms achieving jump heights as high as 30 meters \cite{Hawkes2022, Kovac2008, Zaitsev2015}. Jumping offers the advantage of extended ground contact time, compared to the rapid stance time of hopping platforms, which facilitates self-righting \cite{Kovac2009, Kovac2009a, Fiorini1999, Zhoa2010} and traversal \cite{Zhao2011, Zhao2013} (via legs \cite{Lambrecht2005, Johnson2013, Brill2015, Jung2016} or wheels \cite{Fiorini2003, Stoeter2005, Ho2007}). However, the principles used to achieve high specific energy in jumping platforms can be applied to hopping systems. The design of the hopping leg and its connection to energy storage is crucial for efficient energy transfer. Jumping systems have demonstrated various leg mechanisms, such as linkage \cite{Bonsignori2009, Zhao2013, Plecnik2017, Fiorini2003}, linear \cite{Zhao2009, Scarfogliero2007, Ho2012, Aguilar2016}, buckling \cite{Yamada2007, Yamada2008, Yamada2010}, and spring-based legs \cite{Hawkes2022}. Of these, linkage-based legs may present challenges for hopping systems due to their added complexity and potential failure points from repeated impacts. Jumping platforms have also employed a wide range of additional locomotion strategies, from gliding \cite{Woodward2011, Kovac2011, Woodward2014, Desbiens2013, Desbiens2014, Beck2017, Woodward2019} to flying \cite{Zhu2022}.

Researchers have been studying hopping robots since they were first developed \cite{Raibert1984, Raibert1984a, Sayyad2007}. However, due to their design and control complexity necessitated by the requirement to maintain orientation and input energy over the extremely short time periods and complex ground interactions, only a handful of hopping robots have been developed. For hopping platforms energy can be added in the stance and/or aerial phases to alter or maintain a hoping height. The Salto/Salto 1-p and LEAP are two platforms which hop with stance based energy methods \cite{kulic_untethered_2017, Haldane2016,Plecnik2017,Haldane2017, yim_precision_2018,yim_drift-free_2019,Yim2020}. Alternatively, energy added in the aerial phase has been demonstrated by rotor-based platforms including PogoDrone, PogoX and Hopcopter \cite{Zhu2022, wang_terrestrial_2023,kang_fast_2024, bai_agile_2024}. This style of hopping platform is also more efficient than flight, as hopping is possible with a thrust to weight ratio (TWR) $<$ 1 while flight requires a TWR$\geq1$ \cite{bai_agile_2024}.

Hopping platforms typically maintain stable heights with consistent vertical displacement, though there are exceptions such as hopping onto higher surfaces \cite{yim_precision_2018, wang_terrestrial_2023}, leaping off walls or platforms \cite{Haldane2016}, or modifying energy between hops \cite{Okubo1996, Scarfogliero2006}. Of these, the mechanisms behind energy modification have not been thoroughly investigated, and this is the focus of our work. Specifically, we are interested in energy modification while maintaining a constant actuator power output as opposed to short periods of high output. This results in lighter actuators and higher performance, as seen in a key performance metric for vertical jumping and hopping, $height_{jump} = energy_{jump}/(mass*gravity)$; which is closely tied to mass. This is particularly relevant as hopping robots vary greatly in weight, from Salto's 100 g to the Cargo robot's 150 kg \cite{guenther_energy-efficient_2017}. Additionally, through energy accumulation, this concept can allow the robot to achieve greater hopping heights over multiple cycles. This is inspired by the desert kangaroo rat (Dipodomys deserti), which increase their locomotion energy (LE, defined as the total kinetic, gravitational, and elastic energies) over successive hops by accumulating LE in both kinetic energy and elastic tissues, all while maintaining a constant metabolic rate \cite{Christensen2022}. This approach of accumulating and releasing energy without increasing actuator power provides a model for efficient hopping. Our goal in this work is to explore efficient design strategies for hopping platforms and to evaluate energy modification at hopping heights up to $\sim4m$ ($\sim$2.4x the current state-of-the-art).

\begin{figure*}[tbp!]
\centerline{\includegraphics[width=1.0\textwidth]{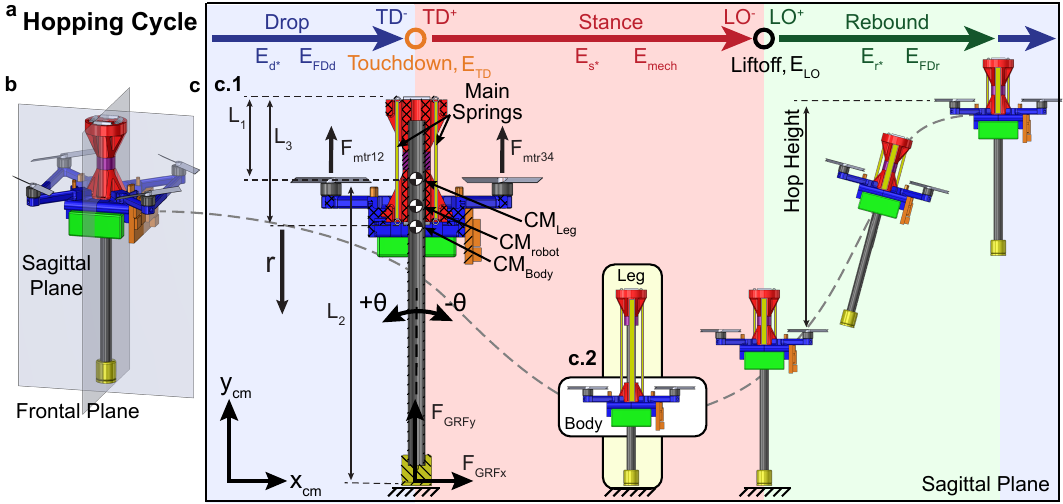}}
\caption{a) Hopping cycle: Phases (drop, stance, and rebound) and transitions (touchdown, liftoff); with locomotion energy (LE) modifications: Control inputs ($E_{d^*}$, $E{s^*}$, $E_{r^*}$), and losses ($E_{FDd}$, $E_{TD}$, $E_{mech}$, $E_{LO}$, $E_{FDr}$). b) Symmetry plane of the MultiMo-MHR. c) CAD model representation of the hopping cycle. c.1) Section view of the CAD model showing internal connections, 4 degrees-of-freedom, and dynamic parameters. c.2) CAD model at max spring extension during the stance phase showing the two bodies (Body, Leg).}
\label{fig:hopCycle}
\end{figure*}

\section{Dynamic Model} 
Locomotion tends to require cyclic motions in which the LE is modified over the course of the cycle. Figure \ref{fig:hopCycle}a shows the hop cycle divided into three phases, including drop, stance, and rebound, and two transitions, including touchdown (TD) and liftoff (LO). Each phase has the potential for a control input to either add or remove LE, whereas both phases and transitions have characteristic losses of LE. This results in eight distinct energy modifications with the total change in LE (apex to apex) as, $m_T g h_{d} (\delta_{rd} - 1) = E_{d^*} + E_{s^*} + E_{r^*} - E_{FDd} - E_{TD} - E_{mech} - E_{LO} - E_{FDr}$; where the total mass, $m_T$, gravity, $g$, drop height, $h_d$, rebound height, $h_r$, height ratio, $\delta_{rd}=h_r/h_d$, represent the total change in energy. The control input energies $E_{d^*}$ (drop), $E_{s^*}$ (stance), and $E_{r^*}$ (rebound), and the energy losses from drag, $E_{FDd}$ (drop) and $E_{FDr}$ (rebound), touchdown, $E_{TD}$, mechanical, $E_{mech}$, and liftoff, $E_{LO}$, are represented accordingly. Given the new class of rotor-based hopping robots, the potential to add energy in rebound ($E_{r^*}$) with a thrust input, $F_{r^*}$, has been facilitated.

To explore hopping with an average rebound control input thrust-to-weight ratio (TWR), $\alpha_{r^*} = F_{r^*}/(m_T g)$, a dynamic model is developed. The model consists of two masses (body, $m_B$, and leg ,$m_L$) connected by a prismatic joint along the long axis of the leg. The losses, measured from the prototype as phase and transition efficiencies, are scaled accordingly: drag ($\eta_{FDd}$,$\eta_{FDr}$), scaled by the total mass $m^{2/3}$, touchdown ($\eta_{TD}$), scaled by the body mass ratio $m_B/m_T$, and mechanical and liftoff ($\eta_{mech}$, $\eta_{LO}$), scaled with fit energy factors. Due to the robot's symmetry about the frontal and sagittal planes (Fig. \ref{fig:hopCycle}b), it can be modeled in the sagittal plane (Fig. \ref{fig:hopCycle}c) resulting in four degrees-of-freedom (DoF) including: the center-of-mass (CM) position [$x_{cm}$, $y_{cm}$], orientation, $\theta$, and spring stretch, $r$; state vector , $q = [x_{cm}, y_{cm}, r, \theta]$. The dynamics are then determined using the Lagrange method as,
\begin{align}
    M &\Ddot{q} + Cg(\dot{q},q) = \tau \label{eq:model}\\
    M &= diag
    \begin{bmatrix}
        m_T \\
        m_T \\
        (m_B m_L)/m_T \\
        c_1/m_T + I_{cmB} + I_{cmL} 
    \end{bmatrix}^T \nonumber\\
    Cg &=
    \begin{bmatrix}
        0 \\
        g*m_T \\
        c_2/m_T \\ 
        m_B m_L \dot{\theta} \dot{r}(2 L_3 - 2 L_1 + 2 r)/m_T 
    \end{bmatrix} \nonumber\\
        \tau &= 
    \begin{bmatrix}
        -U_1 sin(\theta) \\
         U_1 cos(\theta) \\
         -(U_1 m_{L})/m_{T} \\
         -U_2 d_{mtrx}
     \end{bmatrix} 
     +
     \begin{bmatrix}
        F_{GRFx} \\
        F_{GRFy}  \\ 
        c_{3} \\ 
        c_{4} c_{5}
     \end{bmatrix} 
     + F_{drag} + F_{sd} \nonumber
\end{align}
\begin{align}
    c_1 &= m_B m_L (L_1^2 - 2 L_1 L_3 - 2 L_1 r + L_3^2 + 2 L_3 r + r^2) \nonumber\\
    c_2 &= m_T (K_s (r_{sp} + r) + K_{bl} r) + m_B m_L\dot{\theta}^2(L_1 - L_3 - r) \nonumber\\
    c_{3} &= -m_{B} (F_{GRFx} sin(\theta) - F_{GRFy} cos(\theta))/m_{T} \nonumber\\
    c_{4} &= c_{10} (L_2 m_{T} + m_{B}(L_1 - L_3 - r))/m_{T} \nonumber\\
    c_{5} &= F_{GRFy}  sin(\theta) + F_{GRFx} cos(\theta) \nonumber
\end{align}
where, the body mass, $m_B$, leg  mass, $m_{L}$, total mass, $m_T=m_B+m_L$, design lengths, $L_1, L_2, L_3$, (see Fig. \ref{fig:hopCycle}c.1), body CM rotational inertia, $I_{cmB}$, leg CM rotational inertia, $I_{cmL}$, main power spring constant, $K_s$, main power spring pre-stretch, $r_{sp}$, and body-leg contact spring constant, $K_{bl}$, make up the dynamic model and are given in Table \ref{tab:robot_parm} for our system. Several forces and torques are exerted on the robot throughout hopping locomotion phases including, motor thrust, ground reaction forces, drag, and spring damping. First, the motor thrusts in the body frame ($F_{mtr12}, F_{mtr34}$) can be converted into the control inputs $U_1 = F_{mtr12}+F_{mtr34}, U_2 = F_{mtr12}-F_{mtr34}$. The vertical and horizontal ground contact forces ($F_{GRFx}, F_{GRFy}$) in the world frame arise as the robot's foot makes contact with the ground. The final two forces are the drag force, $F_{drag}$, which is assumed to act at the CM of the system and the body-leg contact spring damping force, $F_{sd}$, assumed to act in the $r$-direction.

\begin{figure}[tbp]
\centerline{\includegraphics[width=0.5\textwidth]{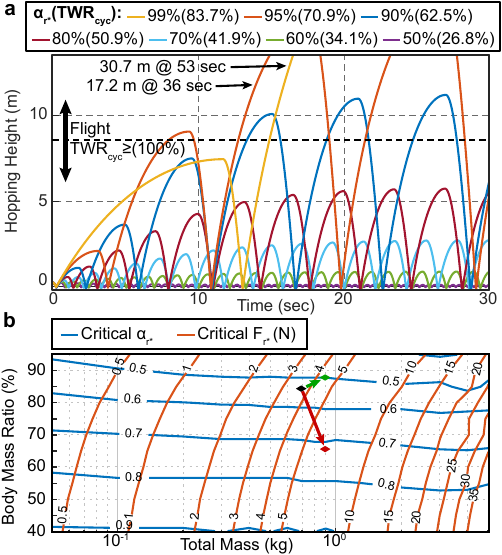}}
\caption{Limited rebound control input TWR, $\alpha_r^* \leq 99\%$. a) Energy accumulation in the MultiMo-MHR with varying $\alpha_{r^*}$, and corresponding cycle averaged TWR, TWR$_{cyc}$, for constant hopping height. b) Critical $\alpha_{r^*}$ ($\alpha_{r^*_{cr}}$, blue lines, minimum TWR for hopping at 0.25 m) and corresponding critical thrust, $F_{r^*_{cr}} = \alpha_{r^*_{cr}} m_T g$ (red lines) for continuous hopping. The MultiMo-MHR (black dot) with 200 grams added to either the body (green dot) or leg (red dot) is also shown.}
\label{fig:concept}
\end{figure}

\begin{figure}[tbp]
\centerline{\includegraphics[width=0.5\textwidth]{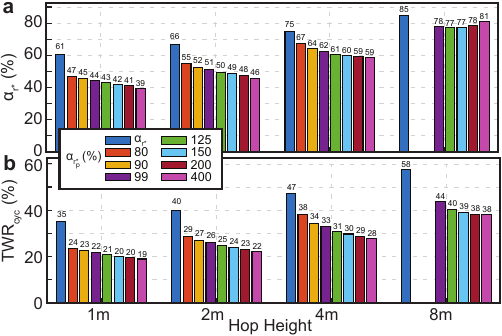}}
\caption{Hopping at various prescribed rebound control input TWRs ($\alpha_{r^*_p}$ = [$\alpha_{r^*}$,80,90,99,125,150,200,400]\% ); where each is maintained for the period necessary to reach the desired hop heights [1,2,4,8] m. The resulting $\alpha_{r^*}$ (a) and TWR$_{cyc}$ (b) are presented.}
\label{fig:concept_HighTWR}
\end{figure}

\subsection{Limited Rebound Control Input TWR, $\alpha_{r^*} \leq 99\%$}
Figure \ref{fig:concept}a, shows the maximum performance of the prototype at varying $\alpha_{r^*}$ from 50\% to 99\% with the associated cycle average TWR, TWR$_{cyc}$, representing the equivalent continuous TWR. As compared to continuous flight, requiring TWR$_{cyc} \geq 100\%$, the prototype can sustain hopping at an $\alpha_{r^*} = 50\%$ and TWR$_{cyc} = 26.8\%$ and reach a maximum height of 30.7 m at an $\alpha_{r^*} = 99\%$ and TWR$_{cyc} = 83.7\%$. In all cases, TWR$_{cyc} < 100\%$ suggests that the hopping robot would always be more efficient than continuous flight, resulting in longer operation times. However, assuming the prototype's maximum $\alpha_{r^*}$ is greater than that necessary to reach the desired height, the constant $\alpha_{r^*}$ can be modified to a prescribed $\alpha_{r^*}$ ($\alpha_{r^*_p}$) applied for a fixed period of time during the rebound phase. This results in differing TWR$_{cyc}$ values and will be discussed in the following section. 

Figure \ref{fig:concept}b, presents the critical $\alpha_{r^*}$ ($\alpha_{r^*_{cr}}$, blue lines), defined as the minimum TWR the robot must have to sustain hopping at 0.25 m, and the resulting critical thrust (red lines), $F_{r^*_{cr}}=\alpha_{r^*_{cr}} m_T g$. The figure covers three orders-of-magnitude in robot mass, [0.05,5] kg, and body mass percentages from 40\% to 95\%, thus encompassing the range of current hopping robots. It demonstrates that $\alpha_{r^*_{cr}}$ is more sensitive to the body mass ratio, $m_B/m_T$, and less sensitive to the total robot mass. The critical thrust, $F_{r^*_{cr}}$ Fig. \ref{fig:concept}b (red lines), also suggests that adding mass to the robot body has a significantly lower impact on the required thrust magnitude as compared to adding mass to the leg. As an example in Fig. \ref{fig:concept}b, the MultiMo-MHR is represented by the black dot; where, 200 g (28.7\% increase) has been added to either the body (green dot) or leg (red dot). The addition of body mass shows a modest increase in required $F_{r^*_{cr}}$ compared to the addition of leg mass. Therefore, larger body masses require proportionally less $F_{r^*_{cr}}$, as a function of the mass, to sustain hopping; freeing up mass for additional components and complexity. 

\subsection{Prescribed Rebound Control Input TWR, $\alpha_{r^*_p}$}
Assuming the hopping robot has a TWR greater than that required to reach a desired height, instead of an applied $\alpha_{r^*}$ over the full rebound phase, the robot can specify a prescribed rebound control input TWR, $\alpha_{r^*_p} \geq \alpha_{r^*}$, for a set period of time during rebound. Figure \ref{fig:concept_HighTWR} shows the resulting $\alpha_{r^*}$ (a) and TWR$_{cyc}$ (b) over $\alpha_{r^*_p}$ = [$\alpha_{r^*}$,80,90,99,125,150,200,400]\% for hopping heights of [1,2,4,8] m. This demonstrates that between 1 and 4 meter hops, operating at higher $\alpha_{r^*_p}$ will result in overall lower $\alpha_{r^*}$ and TWR$_{cyc}$, thus increasing efficiency and operational time. Moreover, comparing a 1 m to 4 m hop, shows that rotor-based hopping robots with greater TWR can operate at higher hopping height with equal efficiency [TWR$_{cyc}(1 \text{ m}, \alpha_{r^*_p}=\alpha_{r^*}) = 35\%$ and TWR$_{cyc}(4 \text{ m}, \alpha_{r^*_p}=90\%) = 34\%$]. At 8 m the $\alpha_{r^*}$ is not monotonically decreasing with increasing $\alpha_{r^*_p}$, indicating that drag is becoming a dominate force. This suggests that as the height increases, the minimum TWR$_{cyc}$ may not be at the maximum $\alpha_{r^*_p}$ available. Therefore, rotor-based hopping robots should, where possible, be designed with a TWR greater than that required for the desired hopping height to improve efficiency; where the majority of the improvement is achieved at modest increases in the TWR.

\begin{table}[tbp!]
\caption{Robot Parameters}
\begin{center}
\begin{tabular}{p{0.045\textwidth}p{0.18\textwidth}p{0.045\textwidth}p{0.045\textwidth}p{0.04\textwidth}}\\[-4ex]
\hline
\textbf{Parm.} & \textbf{Description} & \textbf{R1}  & \textbf{R2}& \textbf{Units} \\[-1ex]
& &  &  &\\
\hline
$m_B$ & Body Mass & 0.588& 0.562 & kg \\
$m_L$ & Leg Mass & 0.109& 0.122 & kg \\
$I_{cmB}$ & Body Rotational Inertia & 0.0012& 0.0012 & kg m$^2$ \\
$I_{cmL}$ & Leg Rotational Inertia & 0.0019& 0.0021 & kg m$^2$ \\
$K_s$ & Main Power Spring Const. & 622& 828 & N/m \\
$K_{lb}$ & Leg-Body Spring Const. & $400 K_s$ & $400 K_s$ & N/m \\
$b_{lb}$ & Leg-Body Damping Coef.& 100 & 100 & N s/m \\
$L_1$ & Leg Top to CM$_L$ & 0.1053 & 0.0795 & m \\
$L_2$ & Leg Bottom to CM$_L$   & 0.2821 & 0.2899 & m \\
$L_3$ & Body Offset from Top Leg & 0.1191 & 0.1175 & m \\
$d_{mtrx}$ & Motor Offset from Body CM & 0.0820 & 0.0820 & m \\
\hline
\end{tabular}
\label{tab:robot_parm}
\end{center}
\end{table}

\section{Robot Design}
\begin{figure}[tbp!]
\centerline{\includegraphics[width=0.5\textwidth]{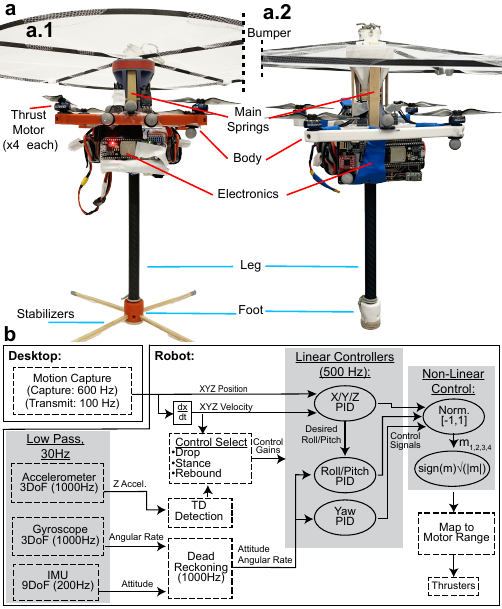}}
\caption{Robot Design: a) MultiMo-MHR with combined monopedal hopping leg and quadrotor for control. a.1) Robot 1 (R1). a.2) Robot 2 (R2).  b) Control scheme.}
\label{fig:robot}
\end{figure}

The MultiMo-MHR (Table \ref{tab:robot_parm}), is designed with a monopedal hopping leg for strong and efficient hopping locomotion, and a quadrotor configuration for energy input and control (Fig. \ref{fig:robot}a); similar to existing robots \cite{Zhu2022, wang_terrestrial_2023,kang_fast_2024, bai_agile_2024}. However, unlike other systems, the body surrounds the leg rather than being above it; minimizing the structural elements required to connect the body and leg. This also maximizes the available leg motion, $d_{spr}$, for a given robot height , $d_{robot}$; whereas, all other hopping robots are limited to a size of $d_{spr} \leq d_{robot}/2 $, the MultiMo-MHR ($d_{robot} = 1.37 d_{spr}$) is limited to a size of $d_{robot} \geq d_{spr}$. Given a specified $d_{robot}$, this design difference allows the required energy ($E_{spr}=1/2 K_{spr} d_{spr}^2$) to be stored with minimum ground reaction forces ($F_{GRF}=K_s d_{spr}$) and therefore minimum loading on the leg structure.

\subsubsection*{\textbf{Energy Storage}}
A major loss of energy in jumping and hopping robots is in the connection between the energy input device and the main elastic energy storage (i.e., spring) device. These mechanisms tend to be complex and experience high forces, which produces significant resistance to the energy release. To remove this impediment to hopping locomotion and maintain high efficiency, the MultiMo-MHR is designed, as with the idealized model, with no connection between the main elastic energy storage device and the energy input device, and therefore these losses are entirely removed. While the MultiMo-MHR utilizes unconnected aerodynamic energy inputs, unconnected inertial energy inputs could be used as well.

Developing a high-specific energy (SE = energy/mass) hopping robot requires consideration of the hopping energy storage medium's SE. Low speed experimental testing of elastomer bands shows a maximum stress (average cross-sectional), strain, and SE of 11.2 MPa, 6.2, and 15800.3 J/kg respectively; where the high speed nature of hopping locomotion and viscoelastic effects of elastomers are expected to increase the stress at lower strains. Therefore, to account for this, and increase robustness and longevity of the elastomers, the MultiMo-MHR is designed to operate with a maximum strain of 2.1. Comparing elastomer bands to optimal helical metallic springs (SE $\sim575$ J/kg) the elastomers have the potential for $\sim27.5x$ greater SE \cite{Woodward2014}. The MultiMo-MHR possesses 6 (Robot 1, R1) or 8 (Robot 2, R2) bands able to store $\sim16.6$ J (4 m hop), which, correlates to an overall SE $\sim24$ J/kg. Hopping robots, with LE input in the rebound phase, can achieve equal hopping heights to their jumping counterparts but with significantly lower SE, and thus ground reaction forces; e.g., the previous MultiMo-Bat at an SE of 78.1 J/kg reaches a height of just 3.6 m \cite{Woodward2019}. 

\subsubsection*{\textbf{Leg Design}}
The monopedal design produces a hopping leg with a single prismatic joint aligned with the locomotion forces. This significantly decouples the friction from the locomotion forces; as the forces are predominantly parallel to the friction direction. Additionally, with the inherent simplicity of the design and primarily compressive forces on the structure, the design produces an extremely strong and robust hopping leg.

Mass scaling is usually assumed to apply to the entire robot; i.e., as body mass increases so too would the leg mass to withstand the increased locomotion forces. However, due to the touchdown in hopping robots, the maximum force experienced by the leg is typically not applied by the body during the stance phase and is instead applied by the foot at touchdown; and is observed in the dynamic model. Additionally, the natural frequency of the foot-ground interface tends to be significantly greater than that of the body-leg. Assuming high damping in the foot-ground interface, the high impact forces will be dissipated before the body can apply any significant forces. Therefore, the determining factor in required leg strength is the leg mass, foot-ground interface stiffness, and touchdown velocity.

Reducing leg mass is generally challenging, and drag couples the total mass to the touchdown velocity at a given height. However, the force during touchdown is highly sensitive to the stiffness of the foot-ground interface, which can be adjusted by modifying the foot contact material. For example, if we assume a touchdown velocity of 7.5 m/s (corresponding to a 4-meter hop) and a maximum foot compression of 5 mm, energy balance calculations predict a peak leg acceleration of 1147 g's (1222 N) at touchdown. In comparison, the maximum body acceleration during stance is 25 g's (146 N); thus, the touchdown forces are $\sim 8x$ greater. At the maximum hop height of 30.7 m ($\alpha_{r^*} = 99\%$, touchdown velocity = 12.8 m/s) which nears the terminal velocity (12.9 m/s), the resulting peak leg g-forces and loads, given a 5 mm foot compression at terminal velocity, are 3393 g's (3615 N); where losses in the dynamic model show these to be over estimates. This therefore decouples the leg mass from the body mass allowing for independent scaling and design.

Given the high compressive strength of carbon fiber tubes, the likely failure would be in buckling. Using the Euler bucking equation, the critical buckling force (CBF) is calculated as, $P_{cr} = \pi^2 E_L I_L/ (K \gamma_{L} L_L)^2 $; where, the young's modulus, $E_L$, area moment of inertia for a tube, $I_L = \pi ((\gamma_{OD} d_{OD})^4-(\gamma_{OD} d_{OD}-2 \gamma_{T} d_T)^4)/64$, outside tube diameter, $d_{OD}$, tube thickness, $d_{T}$, outside diameter scaling factor, $\gamma_{OD}$, thickness scaling factor, $\gamma_{T}$, length factor, $K$, leg length scaling factor, $\gamma_{L}$, and leg length, $L_L$, determine the value. The ratio of the CBF divided by the CBF with scaling factors set to one, $R_{CBF} = P_{cr}/P_{cr}([\gamma_{OD},\gamma_{T},\gamma_{L}]=1)$, presents the overall percent difference in CBF, $P_{cr}$, as the outside diameter leg length, and tube thickness are varied. Using the MultiMo-MHR, the CBF scales, given constant tube thickness ($\gamma_T=1$) and length ($\gamma_L=1$), as follows [$R_{CBF}(\gamma_{OD}=1.25) = 2$, $R_{CBF}(\gamma_{OD}=1.5) = 3.6$, $R_{CBF}(\gamma_{OD}=2) = 8.8$]. The corresponding mass of a tube, $m_{tube} = \pi ((\gamma_{OD} d_{OD}/2)^2-((\gamma_{OD} d_{OD}-2 \gamma_{T} d_T)/2)^2) \gamma_{L} L_L \rho_{cf}$, with material density, $\rho_{cf}$, can also be calculated. As with the CBF, the mass scaling is calculated as, $R_{m} = m_{tube}/m_{tube}([\gamma_{OD},\gamma_{T},\gamma_{L}]=1)$; where the MultiMo-MHR's tube mass scales as follows [$R_m(\gamma_{OD}=1.25) = 1.27$, $R_m(\gamma_{OD}=1.5) = 1.53$, $R_m(\gamma_{OD}=2) = 2.07$]. Therefore, the CBF can be increased significantly without significantly increasing the mass. 

Additionally, as the tube mass represent only $\sim3.2$\% of the total mass, doubling it would result in minor changes to the system's performance with $\sim8.8$x increase in the CBF. Finally, CBF can be increased with no change in mass by reducing the tube thickness as the outside diameter increases; this can be seen by solving the mass scaling equation, $R_m$, for the thickness scaling factor, $\gamma_T$, and substituting it into the CBF scaling equation, $R_{CBF}$ with $R_m = 1$. The CBF of the MultiMo-MHR $P_{cr}$($E_L=150$ GPa, $d_{OD}=16$ mm, $d_T=1$ mm, $K=2$, $L_L=368$ mm) $=3638$ N, demonstrating that the current system can fully withstand ground contact at terminal velocity; where the real CBF will be greater as the mass is distributed over the leg and therefore the loading is not actually concentrated at end of the rod.

\begin{figure}[tbp!]
\centerline{\includegraphics[width=0.5\textwidth]{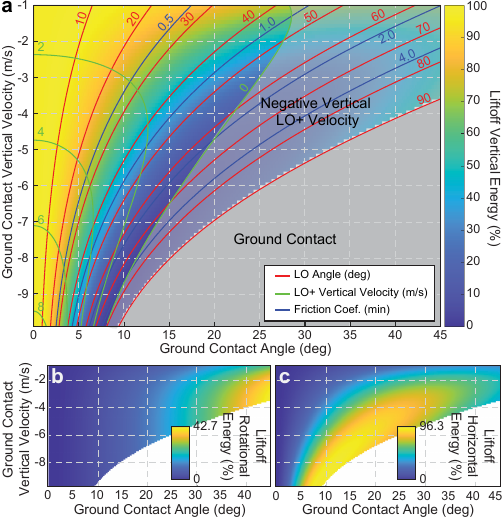}}
\caption{Liftoff (LO) energy percent, as a function of touchdown angle and velocity, divided into translational and rotational components; lost energy not shown. a) LO vertical energy percent. Contour lines show LO angle (red), vertical velocity after LO (LO+, green), and required friction coefficient (blue). b) Rotational energy percent at LO. c) Horizontal energy percent at LO.}
\label{fig:impact}
\end{figure}

\begin{figure}[tbp!]
\centerline{\includegraphics[width=.5\textwidth]{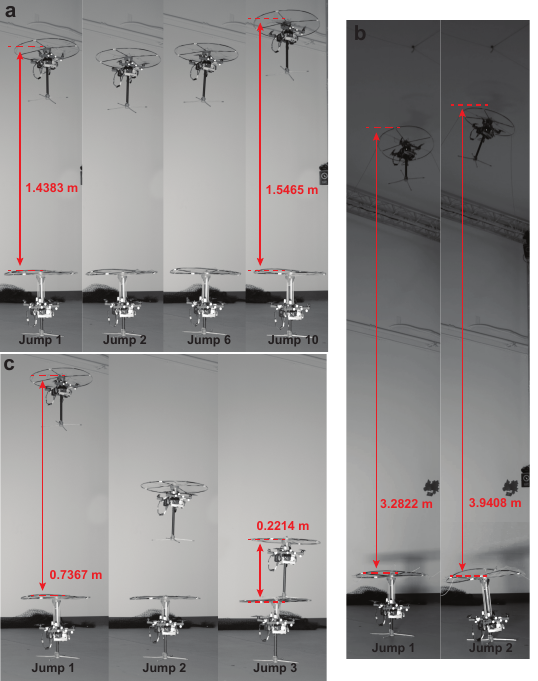}}
\caption{Snapshots of high-speed video (R1 Criteria-2): a) Constant height ($h_d=1.5$ m) $\alpha_{r^*_{p}} = 74.2\%$, b) Increasing height ($h_d=1.5$ m) $\alpha_{r^*_{p}} = 92.2\%$, c) Decreasing height ($h_d=1.5$ m) $\alpha_{r^*_{p}} = 0\%$.}
\label{fig:jump_exp}
\end{figure} 

\begin{figure}[tbp!]
\centerline{\includegraphics[width=.5\textwidth]{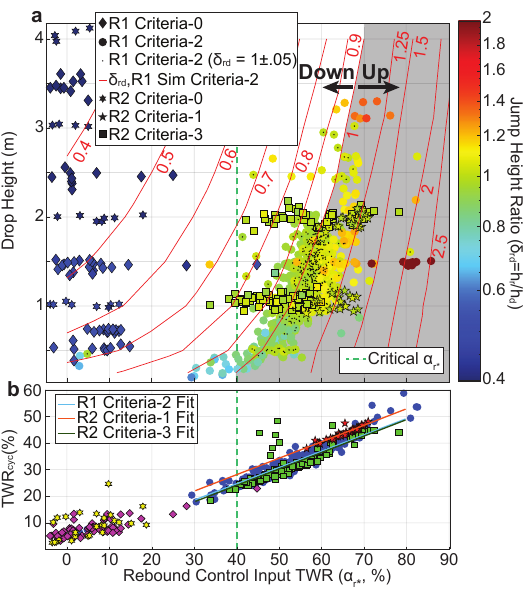}}
\caption{Experimental trials (R1 n = 843 hops, R2 n = 190 hops): a) Hopping height ratio ($\delta_{rd}$) as a function of the rebound control input TWR ($\alpha_{r^*}$) and drop height ($h_d$); where energy can be accumulated (gray region), dissipated (white region), and remain constant ($\delta_{rd}=1$). b) Cycle average TWR (TWR$_{cyc}$) as a function of $\alpha_{r^*}$.}
\label{fig:TWR_exp}
\end{figure}

\subsubsection*{\textbf{Impact Mechanics}} 
As shown in Fig. \ref{fig:impact}, using the dynamic model presented previously, small angular deviations from vertical at touchdown (TD) can lead to significant changes in the liftoff angle, required friction coefficient, and vertical, horizontal, and rotational energies; e.g., a 6 m/s vertical velocity and 10 degree deviation from vertical at ground contact would result in a liftoff angle of 45 degrees and required friction coefficient of 1. Additionally, the system's energy at LO would be divided as follows: vertical axis (35\%), horizontal axis (44\%), rotational axis (4\%), and losses in foot contact (17\%). This poses a significant challenge to increasing the hopping height, requiring the controller to maintain a precise angle and will be discussed in the following section.

\subsubsection*{\textbf{Control}} 
The MulitMo-MHR is based on a standard quadrotor control structure, with alterations given the platform limits TWR $<$ 1. To enable control, seen in Figure \ref{fig:robot}b, two onboard IMUs are utilized: a MPU6050 (6-axis, 1000Hz) for ground detection and a WT901 (9-axis, 200 Hz) for orientation (the MPU6050 gyroscope is used to dead reckon between WT901 readings for 1000 Hz orientation data). A motion capture system (Vicon, Vantage v5, 100Hz) provides the translational information sent to the robot via an RF transceiver (nRF24L01). Finally, the thrust is provided by four quadrotor motors (iFlight XING2 1404 4600 KV with 4030 props), powered all by a 14.7V battery pack (two 7.4v, 2s, 2200 mAh, 35C batteries in series). An important note is our system can only output thrust in a single direction (i.e. cannot reverse motors spin direction); this enables energy input during rebound, but only energy removal in drop (in an ideal case rotor direction would change with velocity direction).  

The fundamental controller is a PID controller with a nonlinear transformation as seen in Figure \ref{fig:robot}b. The control is broken into three components: horizontal [X,Y], orientation [roll, pitch, yaw] and LE input [Z/vertical]. The horizontal PD controllers, active only during rebound, are setup as cascades, generating desired angles for the the roll and pitch controllers. The orientation PD controllers, active during rebound and drop, drive the platform to the desired angles. Lastly, the LE input applies a constant feed-forward thrust in rebound. These combined linear controllers are then sent through a nonlinear transformation; achieved by first saturating and normalizing the linear control output to a [-1,1] range then applying a square root. This accentuates the control for low (i.e near zero) control inputs to maintain the precise angle for TD.

\section{Experimental Results and Discussion}
To validate the model, experimental trials were conducted. Two iterations of the MultiMo-MHR, Fig \ref{fig:robot}a, were utilized, robot 1 (R1, n = 843 hops total) and a slightly updated version robot 2 (R2, n = 190 hops total). R1 uses onboard orientation measurements from the IMU as compared to R2 that uses motion capture data for orientation control; this enhances system stability and enabled the removal of the foot stabilizers. 

The experimental procedure begins by calibrating the sensors with the robot in an upright and stationary position. The robot is then flown up to the desired drop height, ranging from 0.5 to 4 m, and begins hovering. Once stabilized, the hover motor value is recorded for TWR limiting and the drop is initiated; with only the orientation controller active. All controllers are then deactivated at TD. At LO the orientation controller is activated and the horizontal and LE input controllers are active during rebound under specific criteria, including: (Criteria-0, C0) not activated, (Criteria-1, C1) vertical velocity $\geq$ 0 m/s, (Criteria-2, C2) vertical velocity $\geq$ 1 m/s (Fig. \ref{fig:jump_exp} and Supplementary Video 1), (Criteria-3, C3) elapsed time $<$ specified value; where, C0 is designed to determine the cycle efficiency, $\eta_{cyc}$, C1 and C2 are designed to explore maximum hop height at limited TWRs, and C3 is deigned to explore efficiency at higher TWRs. To isolate and explore the LE input control from that required for stabilization, the robot is commanded to hop vertically; minimizing stabilization inputs.

Tests include R1-C0 ($\alpha_{r^*_{p}}$ = 0\%) at heights of [0.5, 1.5, 2.5, 3.5]m and R2-C0 at [1.0, 2.0, 3.0, 4.0]m to determine the system's base efficiency. R2-C1 tests for constant-height hops were initiated at drop heights of [1.0, 2.0]m and $\alpha_{r^*_{p}}$ of [73, 77]\%, for the respective heights. R1-C2 tests were conducted at multiple heights and $\alpha_{r^*_{p}}$ to see constant, accumulating, and decreasing LE over multiple hops. This includes initial heights of [0.5, 1.0, 1.5, 2.0]m with varied $\alpha_{r^*_{p}}$ of [77.5]\%, [70.7, 74.2, 77.5]\%, [74.2, 80.6, 92.2]\% and [74.2]\% for each starting height respectively. Lastly R2-C3 tests for constant-height hops were conducted at heights of [1.0, 2.0]m and $\alpha_{r^*_{p}}$ of [80, 99, 125]\%; where the time is adjusted for each $\alpha_{r^*_{p}}$ to maintain a constant hop height. Each with at least 2-3 hops per trial for higher accumulation/dissipation tests and up to 17 hops per trials at constant hop height.

\begin{table}[tbp!]
\caption{Experimental results at constant hop heights of 1 and 2 m}
\centering 
\begin{tabular}{p{0.05\textwidth}p{0.02\textwidth}p{0.02\textwidth}p{0.06\textwidth}p{0.06\textwidth}p{0.06\textwidth}p{0.06\textwidth}} \\[-4ex]
\hline
\multicolumn{1}{p{0.05\textwidth}}{\textbf{Robot\#-}}& \multicolumn{1}{p{0.02\textwidth}}{\textbf{Hop}} & \multicolumn{1}{p{0.02\textwidth}}{\textbf{$\alpha_{r^*_{p}}$ }} & \multicolumn{1}{p{0.06\textwidth}}{\textbf{$h_d(m)$}}  &\multicolumn{1}{p{0.06\textwidth}}{ \textbf{$\delta_{rd}$}}&\multicolumn{1}{p{0.06\textwidth}}{ \textbf{$\alpha_{r^*}$}} & \multicolumn{1}{p{0.06\textwidth}}{\textbf{TWR$_{cyc}$}} \\[-2ex]
\multicolumn{1}{p{0.05\textwidth}}{\textbf{Criteria\#}}& \multicolumn{1}{p{0.02\textwidth}}{\textbf{Num}} & \multicolumn{1}{p{0.02\textwidth}}{\textbf{(TWR)}} &  & & \multicolumn{1}{p{0.06\textwidth}}{\textbf{(TWR)}} & \\[-2ex]
& &  &  &\\
\hline
R2-C1 & 20& 73   &1.04$\pm$0.08&1.01$\pm$0.03 &63.6$\pm$3.5 &42.8$\pm$2.4\\
R1-C2 & 48& 74.2 &1.03$\pm$0.11&0.98$\pm$0.04 &55.2$\pm$2.4 &34.1$\pm$1.4\\
R1-C2 & 51& 77.5 &1.07$\pm$0.08&1.06$\pm$0.04 &59.6$\pm$2.8 &37.1$\pm$1.9\\
R2-C3 & 21& 80   &1.06$\pm$0.04&1.00$\pm$0.03 &49.5$\pm$4.4 &30.7$\pm$4.0\\
R2-C3 & 20& 99   &1.07$\pm$0.04&1.01$\pm$0.03 &48.6$\pm$5.6 &28.1$\pm$3.0\\
R2-C3 & 33& 125  &1.07$\pm$0.09&1.00$\pm$0.07 &49.6$\pm$6.3 &29.2$\pm$4.6\\
\hline
R2-C1 & 20& 77   &1.96$\pm$0.08&1.00$\pm$0.03 &67.0$\pm$3.1 &44.9$\pm$2.0\\
R1-C2 & 20& 74.2 &1.92$\pm$0.06&0.95$\pm$0.02 &57.9$\pm$2.0 &35.1$\pm$1.0\\
R2-C3 & 15& 80   &2.02$\pm$0.07&1.00$\pm$0.03 &66.7$\pm$4.0 &42.2$\pm$2.3\\
R2-C3 & 16& 99   &1.89$\pm$0.24&1.00$\pm$0.02 &53.1$\pm$6.4 &33.5$\pm$5.9\\
R2-C3 & 15& 125  &2.03$\pm$0.08&1.00$\pm$0.05 &56.2$\pm$8.3 &33.7$\pm$5.6\\
\hline
\end{tabular}
\label{tab:exp_tab}
\end{table}

\subsection{Energy Accumulation and Dissipation} 
Figure \ref{fig:TWR_exp}a shows that the experiential results for R1 (Criteria-0, Criteria-2) are well aligned with the simulation data (R1 Sim Criteria-2, $\delta_{rd}$) with the exception of the low drop height ($< 1 m$) Criteria-0 cases that show increased losses as compared to the simulation; likely due to increased losses in the elastomer due to its non-linear behavior. This demonstrates that the MultiMo-MHR (Criteria-2) is able to accumulate energy at a critical rebound energy input TWR of $\alpha_{r^*} \geq 40\%$; where the up arrow indicates energy accumulation and the down arrow indicates energy dissipation. Assuming the robot begins in the gray region of the plot, it will accumulate energy (increasing the rebound or subsequent drop height) until it reaches the $\delta_{rd}=1$ line, whereas, if it begins in the white region, it will dissipate energy until it reaches the $\delta_{rd}=1$ line or a drop height of zero which indicates cessation of hopping. 

Locomotion requires the robot to both accumulate and dissipate energy. Through having an $\alpha_{r^*}$ greater than the critical value and within the white region of the plot (Fig. \ref{fig:TWR_exp}) the robot is able to dissipate energy such that it will stabilize to a lower hopping height. The maximum reduction in LE per cycle is the hopping cycle efficiency, $\eta_{cyc}$, which can be extracted from the Criteria-0 trials (Fig. \ref{fig:TWR_exp}a) and shows an as expected height dependent $\delta_{rd}$ between 35 (3.5-4 m drop height) and 50 (0.5 m drop height). 

\subsection{Hopping Experimental Efficiency}
The MultiMo-MHRs (R1,R2) are shown to operate at TWR$_{cyc} \leq 100\%$ (Fig. \ref{fig:TWR_exp}b), indicating higher efficiency and therefore longer operational time as compared to conventional flight. This supports our simulation results and the experiential findings from previous work, which indicates that hopping is more efficient than standard flight \cite{bai_agile_2024}.

Comparing the three control criteria C1, C2, and C3 (Fig. \ref{fig:TWR_exp}b, fit lines), we see similar slopes with differing vertical offsets. Confirming the simulation data (Fig. \ref{fig:concept_HighTWR}), when the motors are on for a majority of rebound (C1) we see the largest TWR$_{cyc}$ for a given $\alpha_{r^*}$. Figure \ref{fig:TWR_exp}b (fit lines) shows the cycle average reduces with C2 and is marginally better with C3. This is further observed in Table \ref{tab:exp_tab} with average values for 1 and 2 m hops where $\delta_{rd}\simeq 1$. Comparing R2-C1 ($\alpha_{r^*_{p}} = 73\%$) to R2-C3 ($\alpha_{r^*_{p}} = 125\%$) at an $h_d \simeq 1$ m, we see a reduction in the TWR$_{cyc}$ of $\sim 13\%$; where an $\sim 11\%$ reduction is seen for $h_d \simeq 2$ m. This validates the simulation findings in Figure \ref{fig:concept_HighTWR}, indicating a larger $\alpha_{r^*_{p}}$ used for less of the rebound phase leads to a lower $\alpha_{r^*}$ and TWR$_{cyc}$; indicating increased efficiency.

\section{Summary}
Our results demonstrate a new high-performance hopping robot (MultiMo-MHR) able to hop over $\sim2.5x$ greater than the previous state-of-the-art; reaching heights of nearly 4 meters with the strength to survive impact at terminal velocity or $\geq 30.7$ m. Several control schemes were studied showing higher control input TWRs, over shorter periods of the rebound phase, result in improved overall efficiency. More broadly this work creates a platform with sufficient payload (sensing, communication, computation) and performance to study artificial intelligence, machine learning, and edge computing under highly dynamic locomotion conditions in a miniature robot. Additionally, as high-speed terrestrial locomotion dynamics converge to hopping type behaviors, the concepts developed here, and those that could be developed with the platform, have the potential to be broadly applicable to advancing terrestrial robotics in general.

\bibliographystyle{IEEEtran.bst}
\bibliography{IEEEabrv,bibitems_v1}

\end{document}